\documentclass[sigconf]{acmart}

\usepackage{bbm}  
\usepackage{adjustbox}
\usepackage{array}
\usepackage{multirow}
\usepackage{colortbl}
\usepackage{ulem}
\usepackage{makecell}
\usepackage{balance}
\usepackage{pifont}

\definecolor{mygray}{gray}{.9}

\AtBeginDocument{%
  \providecommand\BibTeX{{%
    \normalfont B\kern-0.5em{\scshape i\kern-0.25em b}\kern-0.8em\TeX}}}

\copyrightyear{2023}
\acmYear{2023}
\setcopyright{acmlicensed}\acmConference[MM '23]{Proceedings of the 31st
ACM International Conference on Multimedia}{October 29-November 3,
2023}{Ottawa, ON, Canada}
\acmBooktitle{Proceedings of the 31st ACM International Conference on
Multimedia (MM '23), October 29-November 3, 2023, Ottawa, ON, Canada}
\acmPrice{15.00}
\acmDOI{10.1145/3581783.3612107}
\acmISBN{979-8-4007-0108-5/23/10}

\acmSubmissionID{1817}

\begin{document}

\title{Transferring CLIP's Knowledge into Zero-Shot Point Cloud Semantic Segmentation}

\author{Yuanbin Wang}
\affiliation{
  \institution{Institute of Artificial Intelligence, Beihang University}
\country{}
}
\author{Shaofei Huang}
\authornote{Corresponding author.}
\affiliation{
  \institution{Institute of Information Engineering, Chinese Academy of Sciences}
\country{}
}
\affiliation{
  \institution{School of Cyber Security, University of Chinese Academy of Sciences}
\country{}
}
\author{Yulu Gao}
\affiliation{
  \institution{Institute of Artificial Intelligence, Beihang University}
\country{}
}
\author{Zhen Wang}
\affiliation{
  \institution{Didi Chuxing}
\country{}
}
\author{Rui Wang}
\affiliation{
  \institution{Didi Chuxing}
\country{}
}
\author{Kehua Sheng}
\affiliation{
  \institution{Didi Chuxing}
\country{}
}
\author{Bo Zhang}
\affiliation{
  \institution{Didi Chuxing}
\country{}
}
\author{Si Liu}
\affiliation{
  \institution{Institute of Artificial Intelligence, Beihang University}
\country{}
}

\renewcommand{\shortauthors}{Yuanbin Wang et al.}

\begin{abstract}
Traditional 3D segmentation methods can only recognize a fixed range of classes that appear in the training set, which limits their application in real-world scenarios due to the lack of generalization ability.
Large-scale visual-language pre-trained models, such as CLIP, have shown their generalization ability in the zero-shot 2D vision tasks, but are still unable to be applied to 3D semantic segmentation directly.
In this work, we focus on zero-shot point cloud semantic segmentation and propose a simple yet effective baseline to transfer the visual-linguistic knowledge implied in CLIP to point cloud encoder at both feature and output levels.
Both feature-level and output-level alignments are conducted between 2D and 3D encoders for effective knowledge transfer.
Concretely, a Multi-granularity Cross-modal Feature Alignment (MCFA) module is proposed to align 2D and 3D features from global semantic and local position perspectives for feature-level alignment.
For the output level, per-pixel pseudo labels of unseen classes are extracted using the pre-trained CLIP model as supervision for the 3D segmentation model to mimic the behavior of the CLIP image encoder.
Extensive experiments are conducted on two popular benchmarks of point cloud segmentation. 
Our method outperforms significantly previous state-of-the-art methods under zero-shot setting ($+29.2\%$ mIoU on SemanticKITTI and $31.8\%$ mIoU on nuScenes), and further achieves promising results in the annotation-free point cloud semantic segmentation setting, showing its great potential for label-efficient learning.
\end{abstract}





\maketitle

\section{Introduction}
Point cloud semantic segmentation aims at grouping 3D points into meaningful regions with corresponding semantic classes. 
It is a fundamental task in many domains (\textit{e.g.}, autonomous driving~\cite{wu2019ground, cheng2022cenet}, robot navigation~\cite{thrun2006stanley}, virtual reality~\cite{garrido2021point}, \textit{etc}.).
However, current 3D segmentation methods~\cite{tang2020searching, zhu2021cylindrical, sun2019srinet, wu2019ground, wen2020cf} can only recognize a fixed range of classes since they are trained on the closed-set segmentation benchmarks containing limited known classes.
To enable the segmentation model to distinguish more classes, a typical solution is to extend the training classes by increasing more manually annotated data, which is laborious and time-consuming and will further raise the cost.
Besides, when faced with novel classes that are not included in the training set, these methods may still fail to make correct predictions.
The lack of generalization ability to unseen classes hinders the application of point cloud segmentation methods in real-world scenarios.

Recently, large-scale vision-language pre-trained models (\textit{e.g.}, CLIP~\cite{radford2021learning}, ALIGN~\cite{jia2021scaling}, etc) have been widely adopted in the 2D zero-shot and open-vocabulary tasks~\cite{zhou2022extract, zhong2022regionclip, li2022language, xu2022groupvit, zang2022open, gu2021open}.
These models are trained with a large amount of image-text paired data to learn the co-embeddings of visual and linguistic features in a shared space through contrastive learning.
Their ability to associate visual concepts with the corresponding linguistic concepts can be well utilized to enable the generalization of unseen classes in 2D image tasks like 2D semantic segmentation~\cite{xu2022groupvit, zang2022open, gu2021open}.
However, due to the difficulty of collecting paired data of point cloud and text and the non-structureless of point cloud data, building a 3D pre-trained model is much more expensive and challenging than the 2D counterpart.
Even though some works~\cite{zhu2022pointclip, zhang2022pointclip} try to fix this issue by rendering point clouds into depth maps and directly making predictions leveraging the CLIP model similarly, the gap between depth images and RGB images prevents it to make full use of the visual-linguistic associations.
Furthermore, these methods focus on the classification task, where only frame-level understanding of point clouds and texts is required.
How to introduce the CLIP model into 3D semantic segmentation models to make point-level predictions remains a problem that lacks exploration.

\begin{figure}[!htbp]
  \includegraphics[width=\columnwidth]{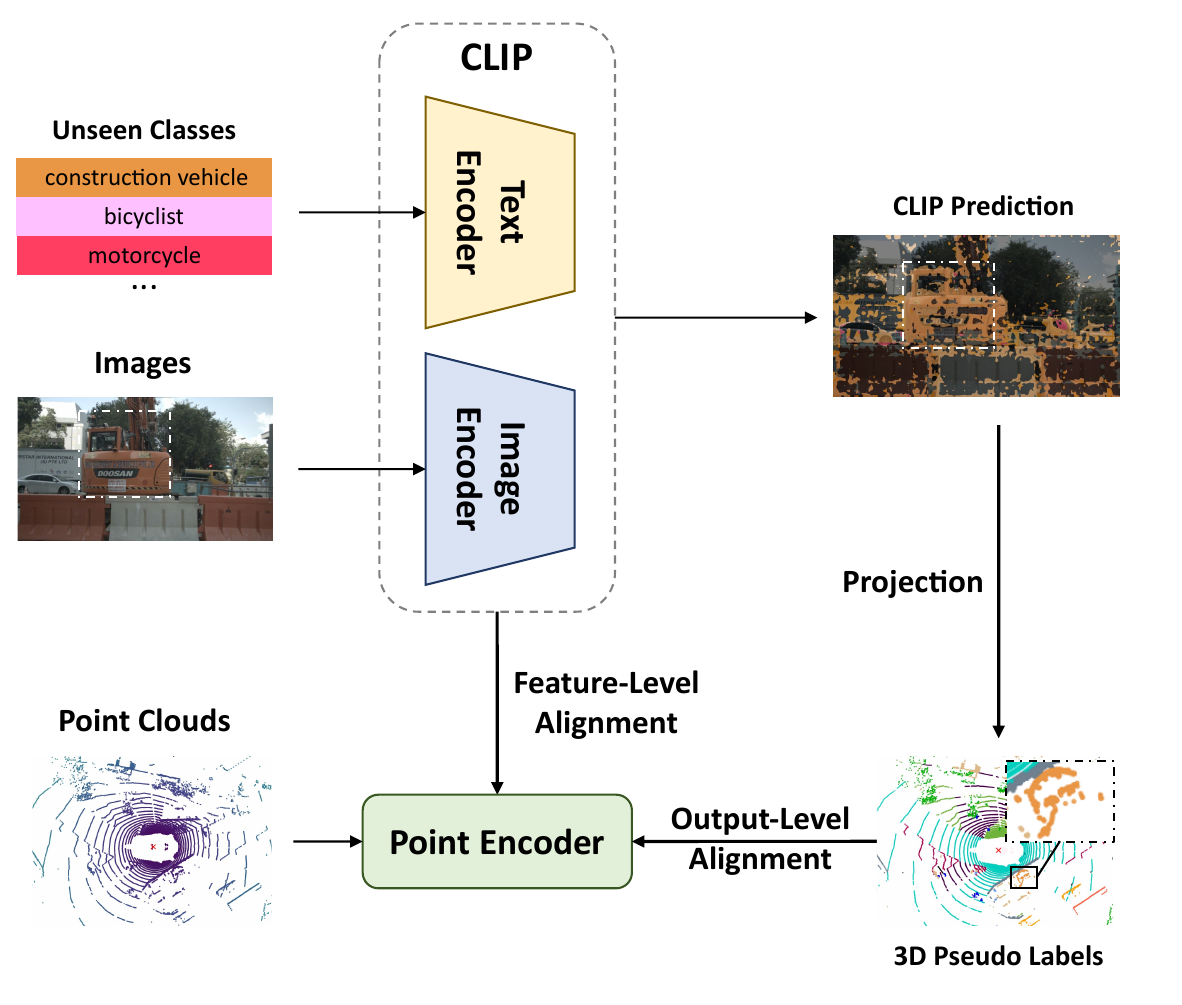}
  \caption{Illustration of our method for zero-shot point cloud semantic segmentation.
  In order to enable 3D point encoder to recognize unseen classes, we leverage the pre-trained vision-language model CLIP and transfer its implied cross-modal knowledge to the 3D point encoder. Both feature-level and output-level alignment between 2D image encoder and 3D point encoder are introduced for effective knowledge transfer.}
  \label{fig:intro}
\end{figure}

In this paper, we focus on the problem of zero-shot point cloud semantic segmentation and propose a simple yet effective baseline to transfer the visual-linguistic cross-modal knowledge implied in CLIP to point cloud segmentation, which is illustrated in Figure~\ref{fig:intro}.
Since there is no explicit correlation between text and point cloud data, visual information serves as a bridge to associate certain text concepts to point cloud regions, so that points belonging to novel classes can be correctly classified.
The 3D point encoder is aligned with the 2D CLIP image encoder at both feature and output levels to sufficiently mimic its behavior.
\textit{For feature level alignment}, we devise a novel Multi-granularity Cross-modal Feature Alignment (MCFA) module that aligns the 2D features extracted by the pre-trained CLIP image encoder and the 3D features extracted by the point encoder from both semantic and position perspectives.
Concretely, prototypes of each class in 2D and 3D representation are extracted by masked average pooling of 2D image features and 3D point features, with 2D and 3D predictions as masks. 
The prototypes corresponding to the same class construct positive samples in contrastive learning for semantic feature alignment. 
On the other hand, 2D and 3D patch features are obtained by average pooling on local patches to form positive and negative samples according to spatial position.
Contrastive loss is calculated between these patch features for position feature alignment.
In this way, the 3D point encoder embeds point cloud features into the same space as features produced by the image and text encoders of CLIP, thus realizing the co-embedding of point cloud, image, and text modalities. 
\textit{For output level alignment}, we utilize the off-the-shelf method MaskCLIP~\cite{zhou2022extract} to assign per-pixel labels of unseen classes for images and generate per-point pseudo labels via projection transformation.
Together with the ground truth labels of seen classes, these pseudo labels of unseen classes are then served as supervision signals to optimize the 3D point encoder.
Through the output level alignment between 2D image encoder and 3D point encoder, the 3D point encoder tends to make similar predictions to the 2D image encoder, which further ensures the effectiveness of cross-modal knowledge transfer.

We conduct extensive experiments on two benchmarks for point cloud segmentation, \textit{i.e.}, nuScenes~\cite{fong2021panoptic} and SemanticKITTI~\cite{behley2019iccv}. 
For the experiments of zero-shot point cloud semantic segmentation, we split all classes into seen and unseen ones and use only seen classes for training following previous works~\cite{michele2021generative, liu2021language}.
Our method outperforms previous methods significantly (\textit{e.g.}, $+29.2\%$ mIoU on SemanticKITTI and $+31.8\%$ mIoU on nuScenes). 
As the pre-trained CLIP is not updated during training, and the network structure of CLIP is not very suitable for segmentation, the performance of the 3D network is limited by the quality of pseudo labels.
Therefore, after training for a few epochs, we apply the self-training strategy and generate pseudo labels of unseen classes with the 3D network itself, bringing a considerable gain of performance to the model.
We further explore the ability of our method in the annotation-free setting, where annotations are not available at all during training, to extract free labels for point cloud segmentation. 
Our method achieves {$18.08\%$} mIoU on the challenging nuScenes benchmark, which further indicates its potential for label-efficient learning.

Our contributions can be summarized as follows:
\begin{itemize}
    \item We propose a framework that transfers the knowledge implied in the pre-trained CLIP model into a 3D segmentation model to facilitate the zero-shot point cloud semantic segmentation. Both feature-level and output-level alignment are conducted between the 2D image encoder and 3D point encoder for effective cross-modal knowledge transfer.
    \item We devise a Multi-granularity Cross-modal Feature Alignment (MCFA) module to enable effective feature-level alignment between the 2D and 3D features from perspectives of global semantic and local position.
    \item Extensive experiments show that our method significantly outperforms previous state-of-the-art methods for zero-shot point cloud semantic segmentation and achieves promising results in the annotation-free setting, which indicates its great potential in the scenario of a label-efficient setting. 
\end{itemize}

\section{Related work}

\subsection{Zero-shot Image Segmentation}
Zero-shot image segmentation~\cite{bucher2019zero, xian2019semantic, zhou2022extract, tian2020cap2seg, gu2020context, ghiasi2022scaling, li2022language, xu2022groupvit} aims at segmenting an image containing classes that are not seen during the training phase. 
The alignment between visual embeddings and text embeddings of categories is of great importance in this task. ZS3Net~\cite{bucher2019zero} follows a generative-based framework and generates fake visual features from semantic word embeddings for pixels belonging to unseen classes. SPNet~\cite{xian2019semantic} adopts an embedding-based framework which projects each pixel of the image features to the text embedding space and assigns per-pixel labels by calculating the feature similarity. 

Recently, large-scale visual-language models like CLIP~\cite{radford2021learning}, have been leveraged in this task due to their great capacity to align between visual and linguistic embeddings. 
LSeg~\cite{li2022language} learns to align pixel-level image features with class embeddings generated by CLIP's text encoder, while OpenSeg~\cite{ghiasi2022scaling} aligns between segment-level image features and image captions through region-word grounding. 
MaskCLIP~\cite{zhou2022extract} investigates the problem of making dense predictions with CLIP by slightly modifying the network structure. 
However, in real scenarios like autonomous driving and indoor navigation, the understanding of 3D scenes is more critical. Therefore, zero-shot learning with point cloud has emerged nowadays.

\subsection{Zero-shot Learning in 3D}
Due to the lack of large-scale paired LiDAR-text datasets, it is more difficult to directly align point and text features using pre-trained models. 
Cheraghian~\textit{et al.}~\cite{cheraghian2019zero} first introduces zero-shot learning for point cloud in the classification task by measuring the cosine similarity with category semantics. Cheraghian~\textit{et al.}~\cite{cheraghian2019mitigating} tackles the hubness problem in 3D zero-shot learning using a novel loss function composed of a regression term and a skewness term. Different from all the above methods, PointCLIP~\cite{zhang2022pointclip} extends CLIP for handling 3D point cloud data by simply projecting each point onto a series of pre-defined image planes to generate scatter depth maps, which doesn't require any training on labeled point cloud.

More recently, some studies~\cite{michele2021generative, liu2021language, chen2022zero, chen2023clip2scene, peng2023openscene} investigate zero-shot learning for point cloud semantic segmentation. They split all classes into seen and unseen ones, and train the model using the annotation of seen classes. Michele~\textit{et al.}~\cite{michele2021generative} adopts a generative-based framework like~\cite{bucher2019zero}, training a feature generator that generates pseudo features for unseen classes. 
Chen~\textit{et al.}~\cite{chen2023clip2scene} establishes point-text correspondence via images and directly align these two modalities. Peng~\textit{et al.}~\cite{peng2023openscene} aligns 3D network with pre-trained 2D open-vocabulary segmentation network by distilling per-point features. Our method performs alignment at both output and feature level to let the 3D network mimic CLIP's image encoder, thus building point-text correspondence.

\begin{figure*}[!htbp]
  \includegraphics[width=\textwidth]{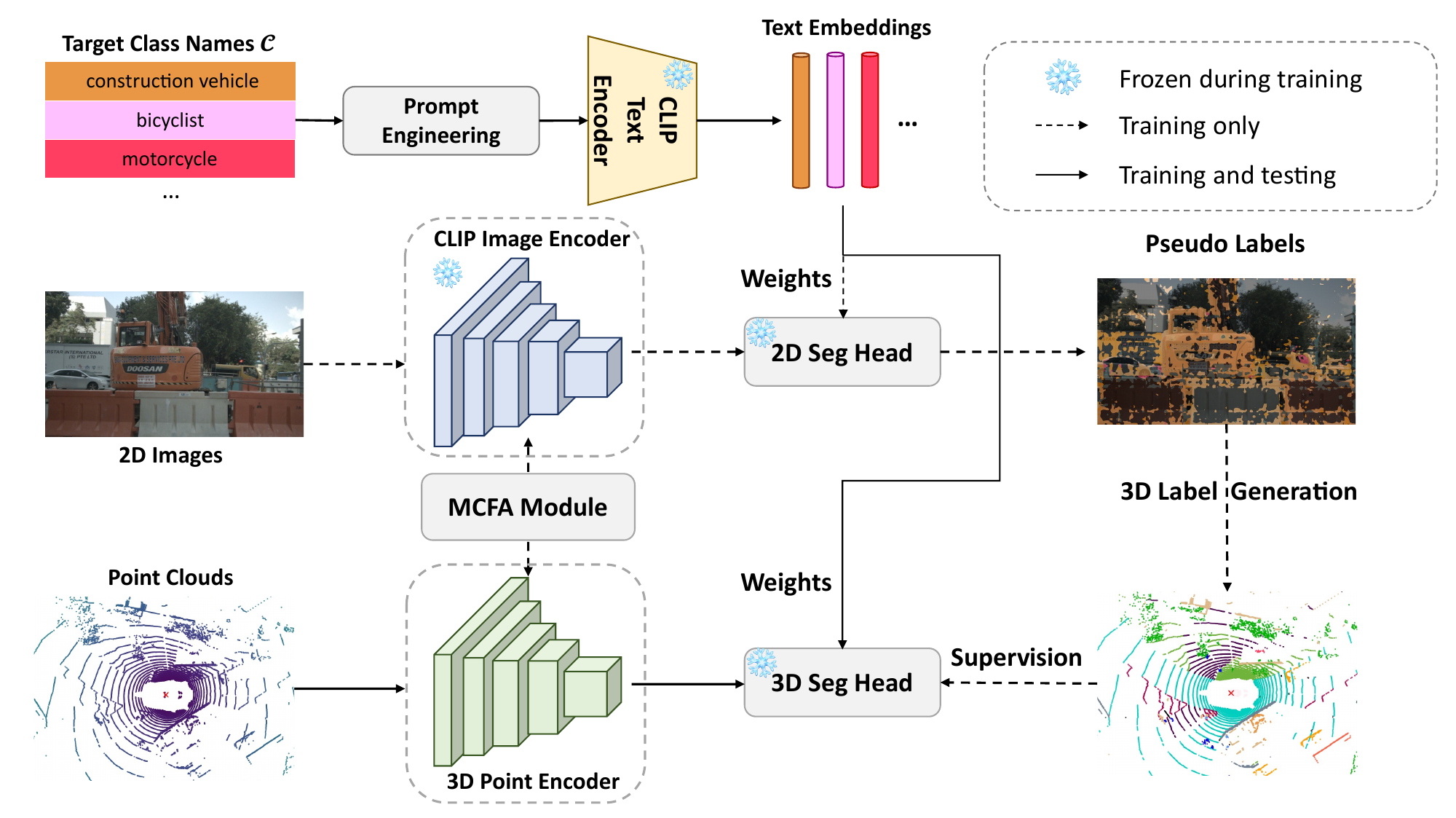}
  \caption{Overall architecture of our proposed method. CLIP's text encoder generates text embeddings after prompt engineering of all class names. CLIP's image encoder and the 3D point encoder produce feature maps of input images and point clouds respectively, with the MCFA module aligning their intermediate features. The text embeddings serve as the weights of both 2D and 3D segmentation heads. Predictions on images are utilized to generate 3D pseudo labels to optimize the 3D point encoder. CLIP's parameters are frozen during training, and only the point clouds are required as input during inference.}
  \label{fig:framework}
\end{figure*}

\subsection{Cross-modal Knowledge Transfer}
Knowledge distillation (KD) is initially proposed for transferring useful information from a large teacher network into a small one. Therefore, the student network is trained to mimic the teacher on the output level~\cite{zhang2019training, hinton2015distilling} or feature level~\cite{wang2022head, mirzadeh2020improved}. With the development of multi-sensor technology, 2D-to-3D distillation methods~\cite{sautier2022image, yan20222dpass, liu20213d, xu2022image2point, chen2023clip2scene, peng2023openscene} are utilized to help models take advantage of data from different modalities, thus improving their performance in 3D tasks. For instance, Xu~\textit{et al.}~\cite{xu2022image2point} proposes to convert pre-trained 2D convolutions into 3D ones via inflation, boosting the performance with minimal finetuning efforts. Yan~\textit{et al.}~\cite{yan20222dpass} transfer knowledge in a multi-scale fusion-to-single manner to better preserve modal-specific information. Sautier~\textit{et al.}~\cite{sautier2022image} apply superpixel-to-superpoint contrastive loss on intermediate feature maps, making the 3D model extract features similar to the 2D pre-trained network on semantically consistent local regions. In our work, both global semantic and local position are leveraged to carry out contrastive learning between CLIP's image encoder and point cloud encoder, transferring image-text alignment into LiDAR-text.

\section{Methods}

\subsection{Background of CLIP}
Contrastive Vision-Language Pretraining (CLIP~\cite{radford2021learning}) aims at learning to project both image and text features in a unified representation space. It consists of an image encoder (ResNet~\cite{he2016deep} or ViT~\cite{dosovitskiy2020image}) and a text encoder (Transformer~\cite{vaswani2017attention}), which are jointly trained with a contrastive loss on abundant image-caption pairs freely collected on the Internet. Matched image-text pairs within a mini-batch are constructed as positive samples, while mismatched pairs are considered negative. 

CLIP has been widely applied to 2D zero-shot classification tasks. 
Class names are placed in a set of pre-defined prompts
and are then fed to the text encoder to generate text embeddings for each class. 
In the meanwhile, the image encoder processes the images to obtain image embeddings. 
Similarities between text embeddings and image embeddings are calculated to predict image labels. 
MaskCLIP~\cite{zhou2022extract} extends CLIP to 2D zero-shot semantic segmentation by modifying the attention pooling layer of CLIP's image encoder without further training.
Instead of aggregating a global representation vector of the whole image, which is actually a spatial weighted-sum of the incoming feature map followed by a linear layer, MaskCLIP applies directly the linear layer to obtain a dense feature map containing rich local information. Followed by an $1\times 1$ convolution layer with text embeddings as weights, the output logits have the same resolution as the feature map, thus performing dense predictions. 

\subsection{Overview}
We first give a detailed explanation of the formulation of the zero-shot point cloud semantic segmentation task.
Given a set of classes to segment $\mathcal{C}$, we divide it into two subsets, including the seen classes set $\mathcal{S}$ and unseen classes set $\mathcal{U}$, where $\mathcal{S} \cap \mathcal{U} = \emptyset$ and  $\mathcal{S} \cup \mathcal{U} = \mathcal{C}$.
In the training phase, only annotations of $\mathcal{S}$ are available to train the 3D segmentation model.
We take a pair of image and point cloud data together with all class names $\mathcal{C}$ as inputs of our model.
However, during inference, regions of both $\mathcal{S}$ and $\mathcal{U}$ are required to be correctly segmented.
Only the point cloud data and class names are required as inputs during inference.

The overall architecture of our proposed zero-shot point cloud segmentation framework is illustrated in Figure~\ref{fig:framework}.
Given a paired image and point cloud data, the 2D feature and 3D feature are extracted by CLIP image encoder and a 3D point encoder respectively.
Text embeddings of all classes $\mathcal{C}$ are generated by prompt engineering and CLIP text encoder successively.
The 2D feature passes through the 2D segmentation head consisting of $1\times 1$ convolution to produce 2D pseudo labels of unseen classes, with text embeddings as convolution kernels.
Intermediate 2D and 3D features are aligned from semantic and spatial perspectives in the Multi-granularity Cross-modal Feature Alignment (MCFA) module.
Finally, the 3D feature is processed by the 3D segmentation head whose weights are also the text embeddings.

\subsection{Feature Extraction}
Given an image $\boldsymbol{I}\in \mathbb{R}^{H\times W\times 3}$ and point cloud data $\boldsymbol{P} \in \mathbb{R}^{N\times 5}$, we first extract the dense 2D feature map $\boldsymbol{F}_{2D}\in \mathbb{R}^{H'\times W'\times C}$ via the image encoder of MaskCLIP and the 3D feature map $\boldsymbol{F}_{3D}\in \mathbb{R}^{N\times C}$ via a 3D point encoder respectively, where $N$ denotes the number of points, $H'$, $W'$ denote the spatial size of feature map and $C$ denotes the channel number.
In order to facilitate the knowledge transfer between 2D and 3D features, we set the architecture of 3D point encoder similar to 2D image encoder 
and add a linear layer on top of the original 3D point encoder to align the feature dimension.
The target class names are inserted into pre-defined templates to generate corresponding prompts, \textit{e.g.}, \textit{there is a $\{\ \cdot\ \}$ in the scene}.
Afterward, we extract the text embeddings of each class $\boldsymbol{Emb}\in \mathbb{R}^{n\times C}$ with CLIP's text encoder, where $n$ is the number of classes containing an extra "background" class. 
Both CLIP text encoder and image encoder are frozen during training.

\subsection{Multi-granularity Cross-modal Feature Alignment}
\label{sec: mcfa}
\begin{figure}[!htbp]
  \includegraphics[width=\columnwidth]{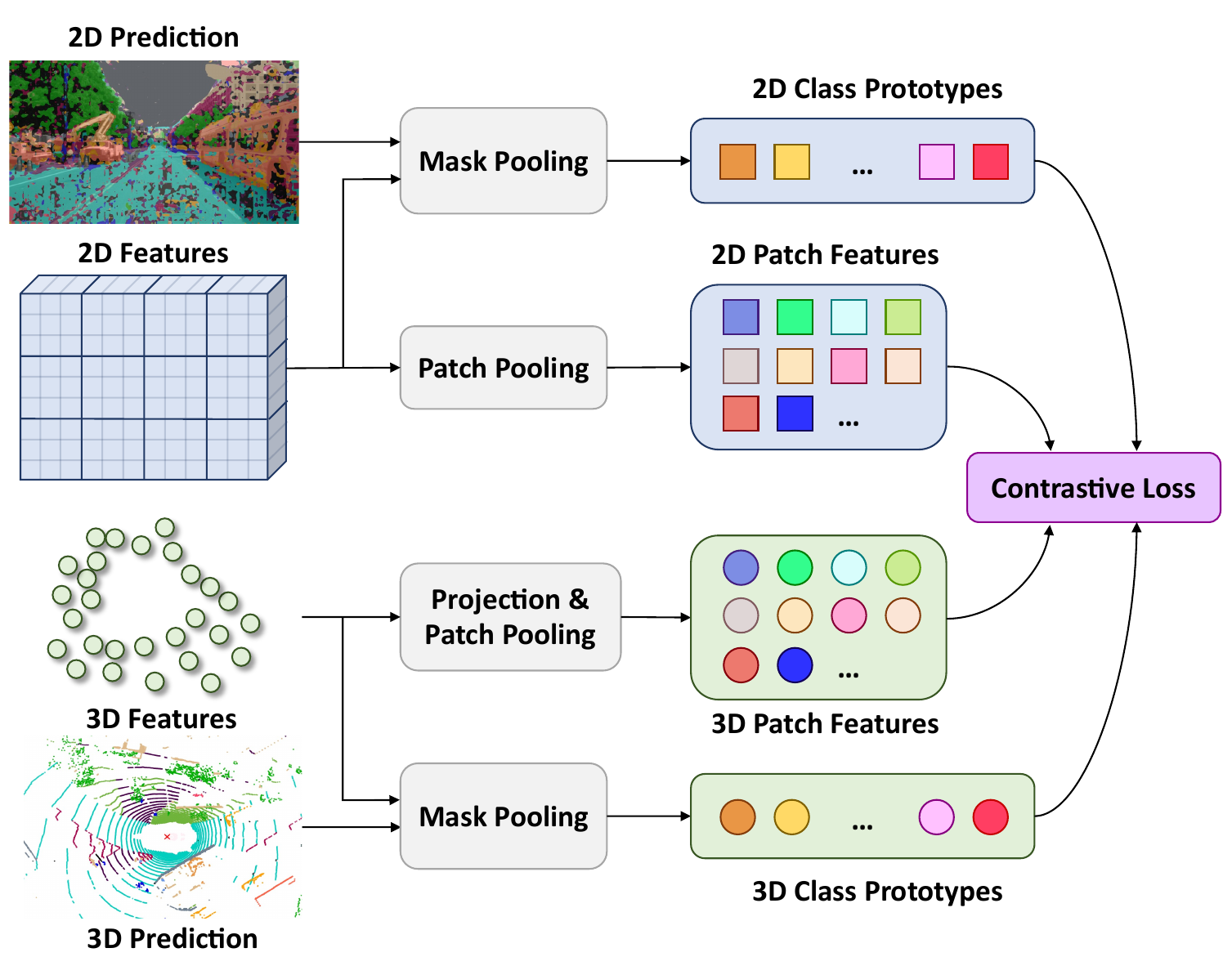}
  \caption{Illustration of the MCFA module. We compute class prototypes and patch features of the two modalities respectively. Contrastive losses are calculated to pull together features of the same semantic class or spatial position while pushing away features from different semantic classes or spatial positions.}
  \label{fig:module}
\end{figure}

Since there is no explicit correlation between the point cloud and text, it is essential to align between image and point embeddings to the shared space first.
In this way, the correlation between the point cloud and text can be established with the bridging of image embeddings.
To this end, we propose a Multi-granularity Cross-modal Feature Alignment (MCFA) module to make 2D and 3D features consistent at the feature level. 
As illustrated in Figure~\ref{fig:module}, two different contrastive losses are calculated between 2D and 3D features to align them from both semantic and spatial perspectives.

\textbf{Class prototype loss.} Given the 2D feature map $\boldsymbol{F}_{2D}^{i}$ and 3D feature map $\boldsymbol{F}_{3D}^{i}$ extracted respectively from the $i$-th stage of 2D image encoder and 3D point encoder, class prototypes of each modality are obtained by masked average pooling with the predictions:
\begin{equation}
    \forall c \in \mathcal{C},\ \ 
    \boldsymbol{g}^{c}_{m} = \frac{\sum\boldsymbol{F}_{m}^{i}\odot\boldsymbol{M}^{c}_{m}}{\sum\boldsymbol{M}^{c}_{m}},
\end{equation}
where $m\in \{2D, 3D\}$ denotes the modality and $\odot$ denotes element-wise product. $\boldsymbol{M}^{c}$ is a binary mask with the same spatial dimension as predictions, where the value is $1$ if this pixel or point is predicted as class $c$. In this way, $\boldsymbol{g}^{c}$ can represent the embedding of class $c$ in images or point clouds. 
During training, prototypes of the same class are considered as positive samples. 
The class prototype loss can be formulated as follow:
\begin{equation}
    \mathcal{L}_{class} = -\sum_{c\in \mathcal{C}}\log\frac{\exp(\langle\boldsymbol{g}^{c}_{2D}, \boldsymbol{g}^{c}_{3D}\rangle/\tau)}{\sum_{c'\neq c}\exp(\langle\boldsymbol{g}^{c}_{2D}, \boldsymbol{g}^{c'}_{3D}\rangle/\tau)},
\end{equation}
where $\langle\cdot, \cdot\rangle$ denotes inner product of two vectors and $\tau>0$ is a temperature.

\textbf{Patch feature loss.} In addition to conducting global contrastive learning on class prototypes, we also leverage the spatial correspondence of images and point clouds to contrast between local patch features, encouraging the 2D image encoder and 3D point encoder to extract similar features for the same spatial position. 
The 3D feature map is firstly projected to the image plane with the relation introduced in Section~\ref{sec:pslabel}, forming a pseudo 2D feature map with the same dimensions as the real 2D feature map provided by 2D image encoder. 
Afterward, we split the two feature maps into $m\times n$ patches and apply global average pooling to each patch, obtaining a series of patch features $\boldsymbol{p}^{i,j}_{2D}$ and $\boldsymbol{p}^{i,j}_{3D}$. 
Features from the same spatial position are considered as positive samples, and the patch feature loss is calculated as follows:
\begin{equation}
    \mathcal{L}_{patch} = -\sum_{i=1}^{m}\sum_{j=1}^{n}\log\frac{\exp(\langle\boldsymbol{p}^{i,j}_{2D}, \boldsymbol{p}^{i,j}_{3D}\rangle/\tau)}{\sum_{(i', j')\neq (i, j)}\exp(\langle\boldsymbol{p}^{i, j}_{2D}, \boldsymbol{p}^{i' j'}_{3D}\rangle/\tau)}.
\end{equation}

The final loss function of MCFA module is summarized as follow:
\begin{equation}
    \mathcal{L}_{cst} = \mathcal{L}_{class} + \mathcal{L}_{patch}.
\end{equation}
By calculating contrastive losses on class prototypes which contain global semantic characteristics and patch features which encode local neighborhood information, the MCFA module builds a tight relation between 2D and 3D features from multiple perspectives, facilitating the knowledge transfer from the pre-trained CLIP image encoder to the 3D point encoder.

\subsection{Output Alignment via 3D Pseudo Label}
\label{sec:pslabel}
To further ensure the effectiveness of cross-modal knowledge transfer, we perform output alignment between 2D and 3D predictions via 3D pseudo labels. First of all, 2D pseudo labels of unseen classes are provided by computing similarities between image features and text embeddings. The 2D pseudo labels are then projected back to 3D space to produce per-point pseudo labels together with the ground truth of seen classes. Finally, segmentation loss is calculated to achieve output-level alignment.

\textbf{2D pseudo label generation.} In order to obtain the supervision signal of unseen classes, we leverage CLIP's generalization ability to provide pseudo labels. First of all, the 2D feature map $\boldsymbol{F}_{2D}$ is up-sampled to the size of the input image, and passes through the 2D classifier, a $1\times 1$ convolution with text embeddings $\boldsymbol{Emb}$ as the kernel. Since only pseudo labels of unseen classes are needed, we select the most probable unseen class or background class as the final result. The process of producing 2D pseudo labels can be summarized as follows:
\begin{gather}
    \boldsymbol{L}_{2D} = \boldsymbol{Emb} \otimes {\rm UpSample}(\boldsymbol{F}_{2D}), \\
    \boldsymbol{\hat{Y}}_{2D} = \mathop{\arg\max}_{\mathcal{U}\cup \{{\rm bg}\}}\ \boldsymbol{L}_{2D},
\end{gather}
where $\boldsymbol{L_{2D}}\in \mathbb{R}^{c\times H\times W}$ is the logit map and $\boldsymbol{\hat{Y}_{2D}}\in \mathbb{R}^{H\times W}$ is the 2D pseudo labels containing only unseen and background classes.

\textbf{3D pseudo label projection.} We complete 3D labels with 2D pseudo labels based on projection principle. Concretely, given the intrinsic matrix $\boldsymbol{K}$ and extrinsic matrix $\boldsymbol{E}$ of the camera, a point $(x, y, z)$ in 3D space can be projected to a pixel $(u, v)$ in the image plan as follows:

\begin{equation}
    \label{eq:proj}
    [u, v, 1]^T = \frac{1}{z}\times \boldsymbol{K} \times \boldsymbol{E} \times [x, y, z, 1]^T.
\end{equation}

Therefore, all points belonging to unseen classes can get a pseudo label by projecting to their corresponding pixels. Together with annotations of seen classes, the generated 3D pseudo labels are utilized to supervise the 3D network. The 3D classifier is a linear layer, whose weights are also text embeddings. 
Thus, point cloud segmentation is achieved by calculating point-text embedding similarity as follows:
\begin{equation}
    \label{eq:seg3d}
    \boldsymbol{P}_{3D} = \mathop{\arg\max}_{\mathcal{C}}\ \boldsymbol{F}_{3D}\cdot \boldsymbol{Emb}^T.
\end{equation}
With the shared weights of 2D and 3D classifiers and pseudo labels provided by CLIP, the 3D network learns to align with CLIP's image encoder on the output level, which further facilitates the unified representation learning for text, image, and point cloud.
We calculate cross-entropy loss and Lovasz loss between the 3D pseudo labels and predictions made by the 3D segmentation head.

\subsection{Training and Inference}
The training scheme of our method contains two phases, \textit{i.e.}, the point encoder is first trained under the guidance of the CLIP model to mimic its behavior at feature and output levels.
After that, it is trained only with pseudo labels provided by itself to further boost the performance.

\textbf{CLIP-guided training.} During the first few epochs, because CLIP is pre-trained while the 3D network is randomly initialized, we utilize CLIP's predictions for unseen classes combined with annotation for seen classes as the segmentation ground truth to supervise the 3D network. We employ cross entropy loss and Lovasz loss as the segmentation loss, and contrastive losses mentioned in Section~\ref{sec: mcfa} are also calculated. The final loss function for CLIP-guided training is expressed as follow:
\begin{equation}
    \mathcal{L}_{s1} = \mathcal{L}_{ce} + \mathcal{L}_{lovasz} + \lambda\mathcal{L}_{cst},
\end{equation}
where $\lambda$ is the weight of the contrastive term.

\textbf{Self-training.} At the end of CLIP-guided training, the performance of the model is limited by the quality of pseudo labels provided by CLIP, which is not updated during training. The predictions of 3D point encoder are more accurate than CLIP. Therefore, in the self-training phase, we utilize the 3D network to produce pseudo labels for unseen classes and supervise itself. Only the cross-entropy segmentation loss is calculated during self-training. 

\textbf{Inference.} During inference, only point clouds and target class names are required as inputs to the model. The process of making per-point predictions is the same as the training phase, which is described in Equation~\ref{eq:seg3d}.

\begin{table*}[htbp]
    \centering
    \caption{Semantic segmentation performance on SemanticKITTI and nuScenes validation set. S denotes seen classes, while U denotes unseen classes. ST means self-training with pseudo labels provided by the 3D network.}
    \label{tab:compare_sota}
    \begin{tabular}{c|cc|c||cccc|cccc}
    \Xhline{1pt}
    \rowcolor{mygray} & & & & \multicolumn{4}{c|}{\bf SemanticKITTI} & \multicolumn{4}{c}{\bf nuScenes} \\
    
    \rowcolor{mygray} \textbf{Setting} & \multicolumn{2}{c|}{\bf Annotation} &  & \multicolumn{3}{c|}{\bf mIoU} & & \multicolumn{3}{c|}{\bf mIoU} & \\

    \rowcolor{mygray} & \textbf{S} & \textbf{U} & \multirow{-3}{*}{\bf Method} & \textbf{S} & \textbf{U} & \multicolumn{1}{c|}{\textbf{All}} & \multirow{-2}{*}{\textbf{hmIoU}} & \textbf{S} & \textbf{U} & \multicolumn{1}{c|}{\textbf{All}} & \multirow{-2}{*}{\textbf{hmIoU}} \\
    
    \hline\hline
    
     & \ding{52} & \ding{52} & SPVCNN~\cite{tang2020searching} & $66.97$ & $60.35$ & $65.58$ & $63.49$ & $83.74$ & $66.38$ & $79.40$ & $74.05$ \\
     
    \multirow{-2}{*}{Supervised} & \ding{52} & & SPVCNN\textsuperscript{*}~\cite{tang2020searching} & $62.31$ & $0$ & $49.19$ & $0$ & $79.10$ & $0$ & $59.33$ & $0$ \\

    \hline
    
    \multirow{5}{*}{Zero-Shot} & \ding{52} & & MaskCLIP-3D+ & $45.88$ & $19.58$ & $40.34$ & $27.44$ & $52.88$ & $35.79$ & $48.61$ & $42.79$ \\
    
    & \ding{52} & & 3DGenZ~\cite{michele2021generative} & $41.40$ & $10.80$ & $35.00$ & $17.10$ & $55.28$ & $20.52$ & $46.59$ & $29.93$ \\
    
    & \ding{52} & & TGP~\cite{chen2022zero} & $54.60$ & $17.30$ & $46.70$ & $26.30$ & - & - & - & - \\

    & \ding{52} & & \textbf{Ours} (w/o ST) & $59.92$ & $26.29$ & $52.84$ & $36.54$ & $77.16$ & $38.72$ & $67.55$ & $51.56$ \\
    
    & \ding{52} & & \textbf{Ours} & $\mathbf{61.31}$ & $\mathbf{46.50}$ & $\mathbf{58.19}$ & $\mathbf{52.89}$ & $\mathbf{79.12}$ & $\mathbf{52.32}$ & $\mathbf{72.42}$ & $\mathbf{62.98}$ \\
    
    \bottomrule
    \end{tabular}
\end{table*}

\section{Experiments}

\subsection{Datasets and Evaluation Metrics}
We conduct our experiments on two popular large-scale benchmarks of point cloud semantic segmentation: SemanticKITTI~\cite{behley2019iccv} and nuScenes~\cite{fong2021panoptic}. 
In zero-shot setting, we follow the previous work~\cite{michele2021generative} to divide all classes of SemanticKITTI into seen and unseen ones, with motorcycle, truck, bicyclist and traffic sign selected as unseen classes. For the nuScenes dataset, we define motorcycle, construction vehicle, traffic cone and trailer as unseen classes, which is similar to SemanticKITTI. 

\textbf{Evaluation metrics.} We evaluate methods using mean intersection over union (mIoU), which is the average IoU over all classes. In addition, since the results may be biased towards seen classes in the zero-shot setting, we report the harmonic mIoU (hmIoU) defined as the harmonic average of the mIoU for seen and unseen classes.

\subsection{Implementation Details}
To obtain comprehensive text embeddings of each class, we follow~\cite{zhou2022extract} to place the class names in $85$ pre-defined prompt templates and take the average of $85$ embeddings produced by the text encoder.
For the 2D inputs, we follow~\cite{yan20222dpass} to randomly crop the input images into a smaller size for faster training. For nuScenes dataset in which images from multiple views are available, we randomly choose one view.

We use ResNet-$50$~\cite{he2016deep} as the 2D image encoder and SPVCNN~\cite{tang2020searching} as the 3D point encoder. We adopt the modification of the attention pooling layer of the image encoder in MaskCLIP to produce dense image features and predictions. We set the hidden dimension of SPVCNN to $128$ for SemanticKITTI and $256$ for nuScenes. According to experimental results, the MCFA module is applied on the 2D and 3D feature maps of the $4$-th stage in each encoder. Before pooling features, a supplementary linear layer converts the 3D features to have the same dimension as the 2D features. In the computation of patch feature loss, we pachify the feature maps with $m=6$ and $n=8$.

Our method is implemented using PyTorch and experiments are conducted on $4$ NVIDIA Tesla A100 GPUs. We apply SGD optimizer with a learning rate of $0.24$, nestrove momentum of $0.9$, weight decay of $10^{-4}$, and batch size of $8$. The hyperparameter $\lambda$ to balance segmentation losses and contrastive losses is set to $0.01$. We train the model for $64$ epochs on SemanticKITTI and $80$ epochs on nuScenes. We set the number of epochs for CLIP-guided training equal to that of self-training.

\begin{figure*}[!htbp]
  \includegraphics[width=\linewidth]{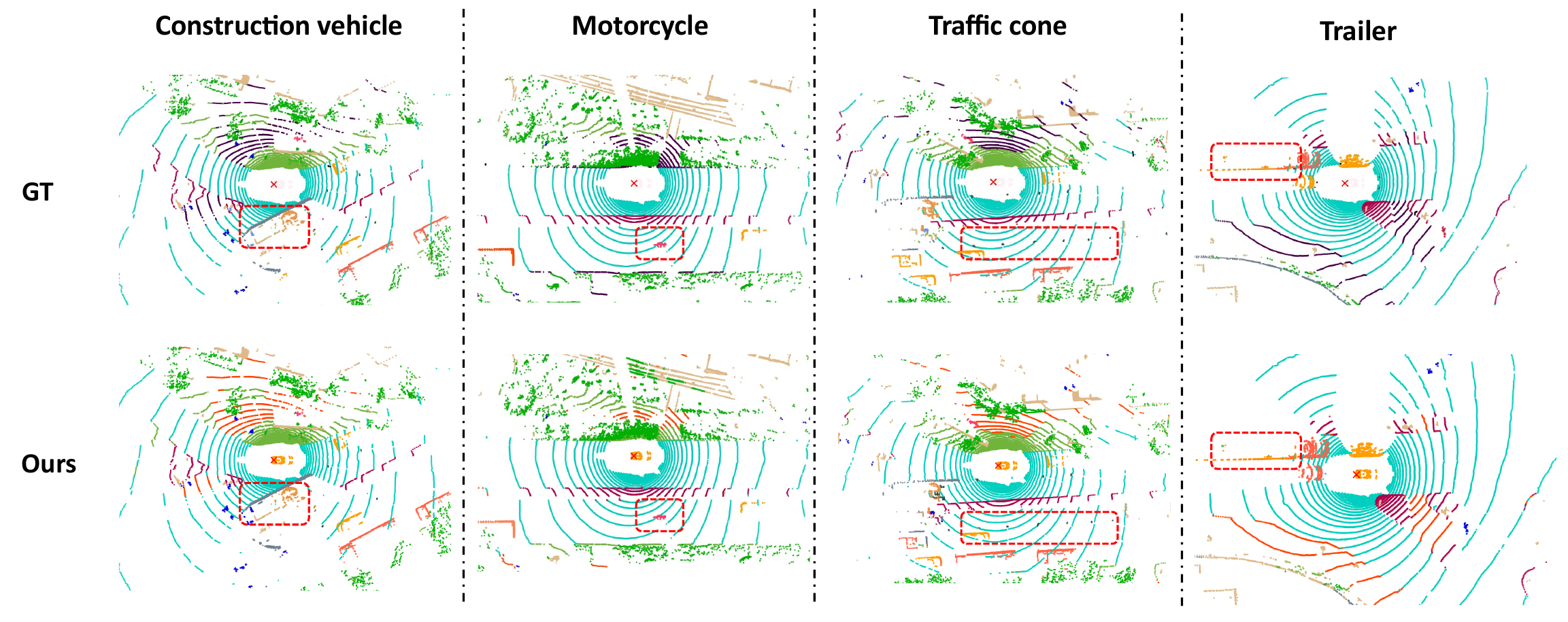}
  \caption{Qualitative results for zero-shot semantic segmentation on nuScenes dataset. Four unseen classes are segmented by our model in a high-quality manner, as shown in \textcolor{red}{red} dashed boxes.}
  \label{fig:vis_zs}
\end{figure*}

\subsection{Zero-Shot Semantic Segmentation}

\textbf{Comparison with state-of-the-art methods.}
Zero-shot semantic segmentation methods train the model only with the labels for a subset of classes (seen classes) and evaluate both seen and unseen classes. In order to demonstrate the effectiveness of our proposed method, we conduct zero-shot experiments on SemanticKITTI and nuScenes benchmarks. The results of the validation sets are reported in Table~\ref{tab:compare_sota}.

To scale our results, we conduct two experiments in a fully supervised manner (the first $2$ rows). In the $1$-st row, the SPVCNN~\cite{tang2020searching} network is trained with annotations of both seen and unseen classes, which is an upper bound for zero-shot methods. In the $2$-nd row, only annotations of seen classes are available, and no adaptation to unseen classes is performed, denoted as SPVCNN\textsuperscript{*}. Therefore, the model can't recognize unseen classes at all, forming the lower bound for zero-shot methods.
In order to show the drawbacks of direct adaptation of 2D methods to 3D tasks and demonstrate the necessity of aligning cross-modal features, we design the MaskCLIP-3D+ model in the $3$-rd row, which simply trains the MaskCLIP+~\cite{zhou2022extract} network on images with MaskCLIP providing pseudo labels for unseen classes, and then obtains point predictions via projection.
Since only the front view images are available in SemanticKITTI, we evaluate MaskCLIP-3D+ only on points that can be projected within the front view image.
We also compare our method with state-of-the-art methods 3DGenZ~\cite{michele2021generative} and TGP~\cite{chen2022zero} for zero-shot point cloud segmentation. 

Among all the above-mentioned methods for zero-shot point cloud segmentation, our method achieves the highest performance under all metrics on both SemanticKITTI and nuScenes benchmarks. Take SemanticKITTI as an example. Without self-training, our method surpasses MaskCLIP-3D+ by $6.71\%$ on mIoU of unseen classes, which indicates the necessity of leveraging 3D point features. Our method surpasses the previous state-of-the-art zero-shot point cloud segmentation method by $8.99\%$ on mIoU for unseen classes and $10.24\%$ on hmIoU. Self-training further boost the performance significantly, with a gain of $20.21\%$ on mIoU for unseen classes and $16.35\%$ on hmIoU. Compared with the fully supervised method in the $1$-st row, our method achieves $75\%$ of its performance in terms of unseen classes without any annotation.


\begin{table}[thbp]
    \centering
    \caption{Ablation studies on SemanticKITTI validation set to verify the effectiveness of the MCFA module.}
    \label{tab: abl_mcfa}
    \adjustbox{max width=1\columnwidth}{
    \begin{tabular}{c|cc||cccc}
        \Xhline{1pt}
        \rowcolor{mygray}  & & & 
        \multicolumn{3}{c}{\bf mIoU} & \multirow{2}{*}{\bf hmIoU} \\ 
        
        \rowcolor{mygray} \multirow{-2}{*}{\bf Baseline} & \multirow{-2}{*}{\bf $\mathcal{L}_{class}$} & \multirow{-2}{*}{\bf $\mathcal{L}_{patch}$} & \textbf{S} & \textbf{U} & \multicolumn{1}{c}{\textbf{All}} & \multirow{-2}{*}{\textbf{hmIoU}} \\

        \hline \hline

        \ding{52} & & & $59.56$ & $21.99$ & $51.65$ & $32.13$ \\

        \ding{52} & \ding{52} & & $59.62$ & $23.81$ & $52.08$ & $34.03$ \\

        \ding{52} & & \ding{52} & $59.19$ & $24.69$ & $51.93$ & $34.85$ \\

        \ding{52} & \ding{52} & \ding{52} & $\mathbf{59.92}$ & $\mathbf{26.29}$ & $\mathbf{52.84}$ & $\mathbf{36.54}$ \\
    \bottomrule
    \end{tabular}
    }
\end{table}

\begin{figure*}[!htbp]
  \includegraphics[width=\linewidth]{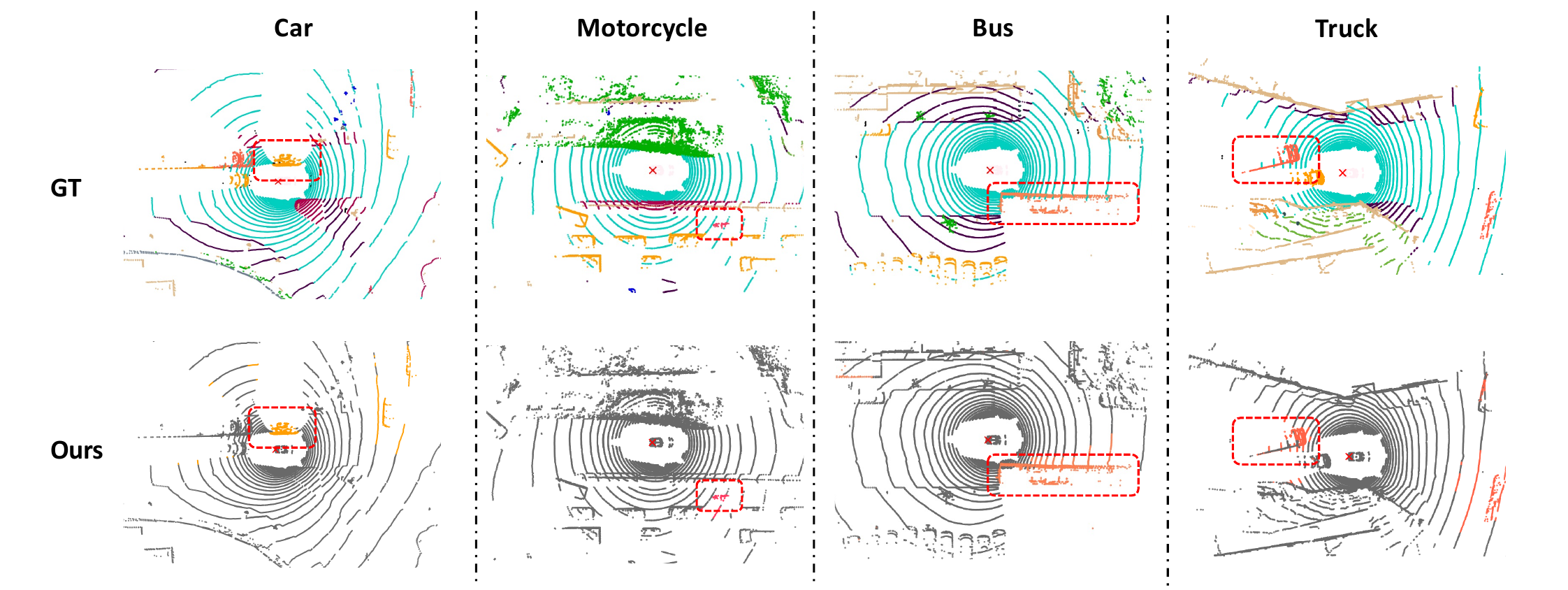}
  \caption{Qualitative results for annotation-free semantic segmentation on nuScenes dataset. Despite the absence of annotations, our model can segment precisely certain classes, (car, bus, \textit{etc.}), as shown in \textcolor{red}{red} dashed boxes. Only the class in question is visualized in our model.}
  \label{fig:vis_af}
\end{figure*}

\textbf{Ablation studies of MCFA module.}
To evaluate the design of the MCFA module, we conduct ablation studies on the validation set of SemanticKITTI. Table~\ref{tab: abl_mcfa} presents the performance when different contrastive losses in the MCFA module are applied. Our baseline only performs output-level alignment, \textit{i.e.}, utilizes pseudo labels of CLIP to supervise the 3D network. In the $2$-nd row, class prototype contrastive loss is calculated to align global class prototypes of images and point clouds, improving the hmIoU by $1.9\%$. Similar results ($+2.72\%$ hmIoU) can be found in the $3$-rd row where patch feature contrastive loss is applied to align 2D image features and 3D point features according to their spatial correspondence. When computing both class prototype and patch feature losses in the $4$-th row, the gains brought by each loss term superimpose, resulting in $4.41\%$ hmIoU improvement compared with the baseline. Through the above ablation studies, each term of the contrastive loss in the MCFA module contributes to the improvement of performance, demonstrating the effectiveness of cross-modal feature alignment from semantic and spatial perspectives.

\textbf{Qualitative results.}
Qualitative results of our method on the nuScenes benchmark are shown in Figure~\ref{fig:vis_zs}. 
For all unseen classes (construction vehicle, motorcycle, traffic cone and trailer), our model can segment them out with precise boundaries. For example, in the $3$-rd column, the traffic cones are so small that can be considered as points in bird's-eye-view. Even though, the model succeeds in predicting every traffic cones in the red box, demonstrating the effectiveness of our model.

\begin{table}[thbp]
    \centering
    \caption{Semantic segmentation performance (mIoU) on SemanticKITTI and nuScnenes validation set under the annotation-free setting.}
    \label{tab: anno_free}
    \adjustbox{max width=1\columnwidth}{
    \begin{tabular}{c|cc||cc}
        \Xhline{1pt}
        \rowcolor{mygray}  {\bf Method} & {\bf Input} & {\bf \#Sweeps} & {\bf SemanticKITTI} & {\bf nuScenes}  \\ 
        \hline \hline
        
        MaskCLIP-3D & C+L & 1 & $8.99$ & $8.33$ \\

        Our baseline & L & 1 & $10.26$ & $14.27$ \\

        {\bf Ours} & L & 1 & $\mathbf{13.17}$ & $\mathbf{18.08}$ \\
        \hline
        CLIP2Scene~\cite{chen2023clip2scene} & L & 3 & - & 20.80  \\
        OpenScene~\cite{peng2023openscene} & C+L & 1 & - & 42.10 \\
    \bottomrule
    \end{tabular}
    }
\end{table}

\subsection{Annotation-Free Semantic Segmentation}

In addition to zero-shot semantic segmentation, we extend our model to annotation-free semantic segmentation. 
We carry out annotation-free experiments on SemanticKITTI and nuScenes datasets, and the results are reported in Table~\ref{tab: anno_free}. 
We compare our method with MaskCLIP-3D, which utilizes MaskCLIP to predict on images and then projects back to 3D space.
Our baseline model without the MCFA module outperforms MaskCLIP-3D, indicating that training with pseudo labels improves the model's performance.
In the $3$-rd row, our model achieves the highest score on two datasets. The MCFA module added in the $3$-rd row brings a performance gain of $2.91\%$ and $3.81\%$ on two datasets, which demonstrates that the MCFA module is useful to align multi-modal features and generalizable to other tasks.
There are some recent works that achieve promising results, but their experimental settings are different from ours. CLIP2Scene~\cite{chen2023clip2scene} employs ViT-L as its image encoder (while we use ResNet-50), and conducts spatial-temporal consistency learning between multiple sweeps (while we only use one), a direct comparison is not possible. OpenScene~\cite{peng2023openscene} distills pre-trained open-vocabulary image segmentation models rather than CLIP and takes both image and point cloud modalities as input during inference, which distinguishes it from CLIP2Scene and our method.

As shown in Figure~\ref{fig:vis_af}, we visualize predictions of certain classes by the model. Compared with the ground truth, even in complicated scenes, our model can perceive some kinds of objects (car, motorcycle, bus, and truck) in spite of the absence of annotations. 
We observe that the classes relatively easier for annotation-free semantic segmentation networks are usually objects with clear spatial shapes, which indicates that 3D geometries may be helpful for this task.
Both the quantitative and qualitative results reveal the great potential of our method toward annotation-free semantic segmentation of point clouds.

\section{Conclusion}

In this paper, we focus on the problem of zero-shot semantic segmentation of point clouds and propose a simple yet effective framework to transfer the visual-linguistic cross-modal knowledge of CLIP to the point cloud segmentation task. The 3D point encoder is aligned with the 2D CLIP image encoder at both feature and output levels to sufficiently mimic its behavior. Extensive experimental results show that our proposed method outperforms significantly previous state-of-the-art works for zero-shot point cloud segmentation, and reveal the potential of our model toward annotation-free semantic segmentation. In the future, we hope to further extend our work to open-vocabulary tasks of 3D vision, supporting any text queries as inputs.

\begin{acks}
This work was supported in part by the National Key R\&D Program of China under Grant 2022ZD0115502, in part by the National Natural Science Foundation of China under Grant 62122010, in part by CCF-DiDi GAIA Collaborative Research Funds for Young Scholars, and in part by CAAI-Huawei MindSpore Open Fund.
\end{acks}

\normalem
\bibliographystyle{ACM-Reference-Format}
\balance
\bibliography{ref}
\begin{figure*}[!htbp]
  \includegraphics[width=\linewidth]{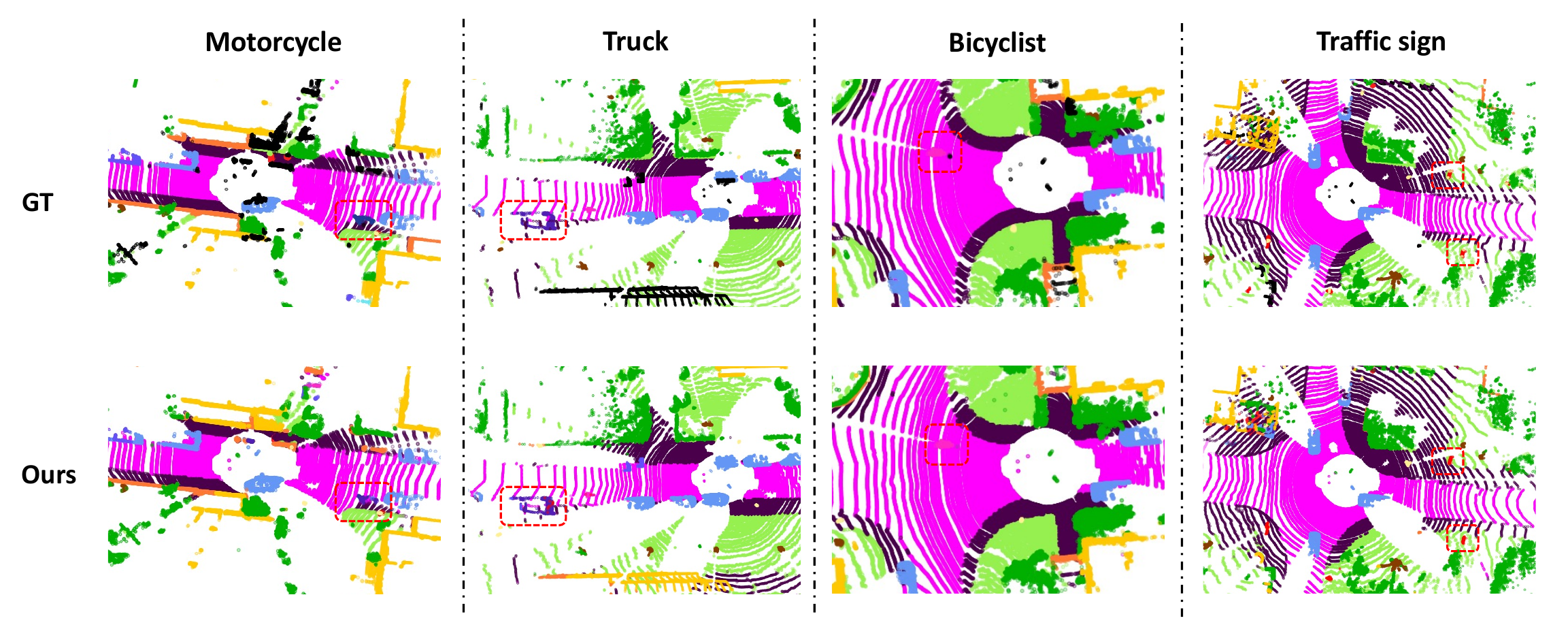}
  \caption{Qualitative results for zero-shot semantic segmentation on SemanticKITTI dataset. Four unseen classes are segmented by our model in a high-quality manner, as shown in \textcolor{red}{red} dashed boxes.}
  \label{fig:vis_zs_sk}
\end{figure*}
\newpage
\appendix

\section{Experimental results}
\begin{table}[htbp]
    \centering
    \caption{Semantic segmentation performance on SemanticKITTI and nuScenes test set.}
    \label{tab: test}
    \adjustbox{max width=1\columnwidth}{
    \begin{tabular}{c||cccc|cccc}
        \Xhline{1pt}
        \rowcolor{mygray} & \multicolumn{4}{c|}{\bf SemanticKITTI} & \multicolumn{4}{c}{\bf nuScenes} \\
        
        \rowcolor{mygray} \textbf{Method} & \multicolumn{3}{c|}{\bf mIoU} & & \multicolumn{3}{c|}{\bf mIoU} & \\
    
        \rowcolor{mygray} & \textbf{S} & \textbf{U} & \multicolumn{1}{c|}{\textbf{All}} & \multirow{-2}{*}{\textbf{hmIoU}} & \textbf{S} & \textbf{U} & \multicolumn{1}{c|}{\textbf{All}} & \multirow{-2}{*}{\textbf{hmIoU}} \\

        \hline \hline
        
        SPVCNN~\cite{tang2020searching} & $67.99$ & $65.18$ & $67.40$ & $66.55$ & $78.72$ & $74.25$ & $77.60$ & $76.42$ \\

        Ours & $64.01$ & $42.10$ & $59.40$ & $50.79$ & $77.11$ & $47.72$ & $69.76$ & $58.95$ \\
    \bottomrule
    \end{tabular}
    }
\end{table}

For further research, we also report our results on the test set of SemanticKITTI and nuScenes in Table~\ref{tab: test}. Since other zero-shot point cloud segmentation methods haven't provided their results on test sets, we list only the fully supervised SPVCNN~\cite{tang2020searching} for comparison.

\section{Qualitative Results}

Qualitative results of our method on the SemanticKITTI~\cite{behley2019iccv} benchmark are shown in Figure~\ref{fig:vis_zs_sk}. For all unseen classes (motorcycle, truck, bicyclist and traffic sign), our model can segment them out with precise boundaries. For instance, in the $1$-st column, the motorcycle is parked on the side of the road. The model correctly segments it even though the background is complicated. Besides, only front view images are available in SemanticKITTI dataset, but unseen objects of the full $360^\circ$ fields of view can be recognized by our model, demonstrating that CLIP's knowledge is successfully transferred to the 3D network.

\end{document}